\documentclass[letterpaper,10pt,twocolumn]{article}
\PassOptionsToPackage{table}{xcolor}
\usepackage[T1]{fontenc}
\usepackage{newtxtext}
\usepackage{newtxmath}
\usepackage[letterpaper,textwidth=7.0in,textheight=9.0in,centering,
            columnsep=0.375in]{geometry}
\usepackage[hyphens]{url}
\usepackage{graphicx}
\usepackage[round]{natbib}
\usepackage{caption}
\usepackage{booktabs}
\usepackage{amsmath}

\usepackage{amssymb}
\usepackage{placeins}
\usepackage{tikz}
\usepackage{xcolor}

\setcitestyle{authoryear,round,aysep={},yysep={,},notesep={, }}

\makeatletter
\def\normalsize{\@setfontsize\normalsize\@xpt{11}}
\def\small{\@setfontsize\small\@ixpt{10}}
\def\footnotesize{\@setfontsize\footnotesize\@ixpt{10}}
\def\scriptsize{\@setfontsize\scriptsize\@viipt{10}}
\def\tiny{\@setfontsize\tiny\@vipt{7}}
\def\large{\@setfontsize\large\@xipt{12}}
\def\Large{\@setfontsize\Large\@xiipt{14}}
\def\LARGE{\@setfontsize\LARGE\@xivpt{16}}
\def\huge{\@setfontsize\huge\@xviipt{20}}
\def\Huge{\@setfontsize\Huge\@xxpt{23}}
\makeatother
\normalsize
\makeatletter
\renewcommand\paragraph{\@startsection{paragraph}{4}{\z@}%
  {-6pt plus -2pt minus -1pt}{-1em}{\normalsize\bfseries}}
\makeatother

\makeatletter
\gdef\@affiliations{}
\newcommand{\affiliations}[1]{\gdef\@affiliations{#1}}
\renewcommand\maketitle{%
  \twocolumn[%
    \begin{@twocolumnfalse}
      \vspace*{0.08in}
      \begin{center}
        {\LARGE\bfseries \@title \par}
        \vskip 0.12in
        {\large\bfseries \@author \par}
        \vskip 0.08in
        {\normalsize \@affiliations \par}
      \end{center}
      \vskip 0.18in
    \end{@twocolumnfalse}%
  ]%
}
\makeatother

\renewenvironment{abstract}{%
  \centerline{\bfseries Abstract}%
  \vspace{0.5ex}%
  \begin{quote}\small}{%
  \end{quote}}

\newcommand{\method}[1]{\textsc{#1}}
\newcommand{\ac}{\method{ap}}
\newcommand{\vlm}{\method{vp}}
\newcommand{\se}{\method{se}}
\newcommand{\sm}{\method{fm}}
\newcommand{\eef}{\method{sp}}
\newcommand{\nostate}{\method{no-state}}

\title{How Should Vision-Language-Action Models Use Proprioceptive State?}
\author{
Yiren Zhao\textsuperscript{\rm 1,3},
Ziyang Chen\textsuperscript{\rm 1,3},
Ziyang Rao\textsuperscript{\rm 1,3},\\
Pengteng Li\textsuperscript{\rm 1,3},
He Zhang\textsuperscript{\rm 1,3},
Weiyu Guo\textsuperscript{\rm 2,3*},
Yandong Guo\textsuperscript{\rm 3},
Rushi Dai\textsuperscript{\rm 1*}
}
\affiliations{
\textsuperscript{\rm 1}The Hong Kong University of Science and Technology (Guangzhou)\\
\textsuperscript{\rm 2}Multimedia Laboratory (MMLab), The Chinese University of Hong Kong\\
\textsuperscript{\rm 3}AI$^2$ Robotics X-Lab
}

\begin{document}
\maketitle
\begingroup
\renewcommand{\thefootnote}{*}
\footnotetext{Corresponding authors.}
\endgroup

\begin{abstract}
Recent Vision–Language–Action (VLA) models almost universally take robot proprioceptive state as input, yet wire it in incompatible ways—serialized into text prompts, projected into the vision–language prefix, or fed directly to the action expert—and almost always as a single current frame. Three questions remain open: (1) whether, and on which tasks, current state actually improves closed-loop control; (2) how much state history helps, and whether its benefit reflects genuine temporal variation rather than added conditioning capacity; and (3) where state should enter the model—the vision–language backbone or the action-generation module. We answer these questions through controlled experiments on a flow-matching VLA, fixing the backbone, training data, action representation, and evaluation protocol throughout. We implement five representative interfaces—discrete state prompt, VLM prefix, action prefix, state expert, and feature modulation—under matched implementation details, and evaluate them on 45 atomic tasks spanning three task families plus 20 composite tasks; we then sweep the state-history length from 1 to 96 frames to examine how historical state information affects model performance. The experiments yield systematic answers to all three questions, distilled into testable design principles for state-aware VLAs.
\end{abstract}

\section{Introduction}

Trained on increasingly large-scale robot demonstrations, recent
Vision-Language-Action (VLA) models can execute hundreds of manipulation
tasks with a single set of weights and generalize across objects and scenes
\citep{intelligence2025pi_,bjorck2025gr00t,chen2026phaser}.
Inspired by biological motor control---animals rely on proprioception to
produce smooth, coordinated, and robust movement---it is natural to expect
the robot's proprioceptive state (joint angles, end-effector pose, gripper
aperture) to play a similar role in a VLA \citep{guo2026brain}: smoother control, more
stable motion phases, and more precise contact-rich manipulation. Yet a gap remains.
Vision and language have been pretrained at scale inside the VLM. Proprioceptive state has no such pretraining. As a result, proprioception remains the least examined and least
understood input in VLAs.

Precisely because no consensus exists, current VLAs use state in strikingly
diverse ways. At the \emph{representation} level, $\pi_{0.5}$
\citep{intelligence2025pi_} discretizes the state into text tokens appended
to the language prompt, whereas OpenVLA-OFT \citep{kim2025fine} and
GR00T~N1 \citep{bjorck2025gr00t} project it continuously into embeddings.
At the \emph{injection} level, some designs feed the state into the VLM,
where it joins images and language in multimodal context modeling, while
others send it directly to the action expert to condition action
generation. At the \emph{temporal} level, the vast majority of methods use
only the current single frame, and few examine the temporal evolution of
state \citep{guo2026brain}. Moreover, these design
differences are
entangled with the backbone,
pretraining, data, and action representation, making them mutually
incomparable. This makes a basic yet unanswered question especially
important: with everything else held fixed, in \emph{what form} should
robot state be represented, \emph{where} should it be injected, and
\emph{how much history} is needed for it to genuinely improve a VLA's
decision making?

\paragraph{Contributions.}
We address these open questions through controlled experimental analysis.
Concretely, we take a representative flow-matching VLA as our testbed and
build a unified experimental framework that fixes the backbone network,
training data, action representation, and evaluation protocol; evaluation
follows a layered protocol built on RoboCasa365
\citep{nasiriany2026robocasa365}, whose atomic tasks we pre-partition by
control semantics into three families (pick-and-place, articulated-object
interaction, and precise actuation), paired with 20 in-distribution
composite tasks. Under this setup, the only thing that varies between systems is how the
state is integrated, so any performance difference can be attributed to
the state design itself. On this controlled basis, we make the following
contributions along the three design axes.

First, for the \emph{representation} question, we compare discrete and
continuous state representations under matched conditions: the same state
is either quantized and serialized into text tokens or projected into
continuous embeddings, and we characterize whether, and on which tasks,
each form of the current state helps.

Second, for the \emph{temporal} question, we sweep the state history from
1 to 96 frames and, with a slot-matched repeat-current control, test
whether history gains require genuine temporal content---how much history
helps, and when more starts to hurt.

\begin{figure*}[t]
\centering
\includegraphics[width=\textwidth]{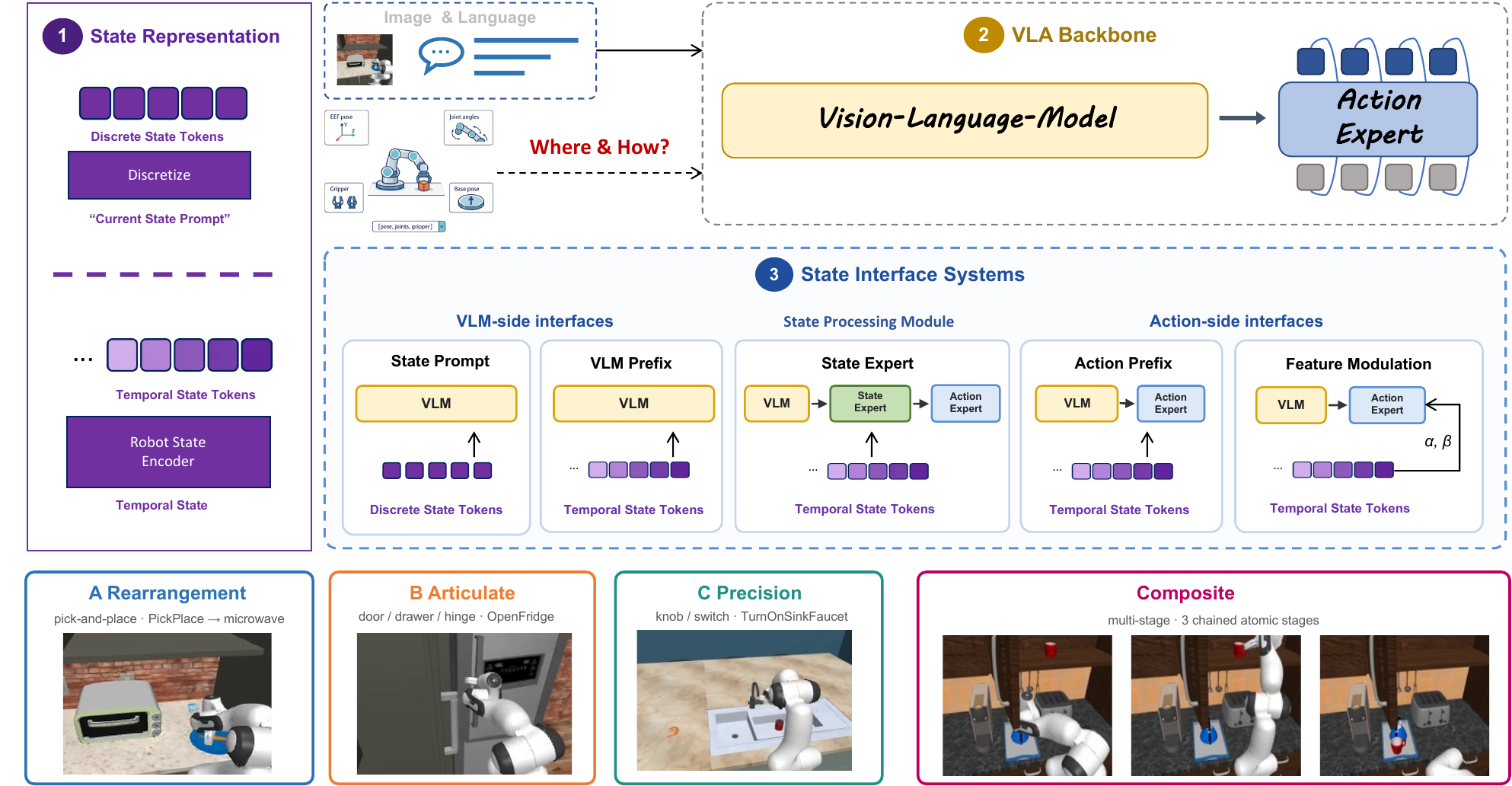}
\caption{Overview of the state design space and evaluation suite.
\textbf{(1) State representation:} robot state is represented either as a
discrete current-state prompt or as continuous temporal tokens.
\textbf{(2) Shared VLA scaffold:} multi-view observations and the task
instruction enter the vision--language backbone, whose context conditions the
action expert. \textbf{(3) State interfaces:} we compare a discrete state
prompt, a continuous VLM prefix, a dedicated state expert, an action prefix,
and feature modulation. The systems are evaluated on rearrangement (A),
articulation (B), precision knob/switch control (C), and multi-stage composite
tasks.}
\label{fig:interfaces}
\end{figure*}

Finally, for the \emph{interface} question, we distill five representative
state interfaces from prior work (Figure~\ref{fig:interfaces})---a
discrete state prompt (\eef{}), a VLM prefix (\vlm{}), an action prefix
(\ac{}), a state expert (\se{}), and feature modulation (\sm{})---implement
them in one scaffold, and compare injecting the same state into the VLM
side versus the action side, showing that the preferred entry point is not
fixed but switches with the temporal budget. Distilling the evidence along
all three axes, we offer testable design guidelines and a reusable
evaluation protocol for future state-aware VLAs.

\section{Related Work}

\paragraph{State-conditioned robot policies.}
Classic visuomotor policies concatenate robot state with visual features
or condition the action decoder on a learned embedding
\citep{DBLP:conf/rss/ZhaoKLF23,chi2025diffusion}. VLA systems inherit
this input but disagree on how it is encoded, where it enters, and what
role it plays
\citep{kim2024openvla,black2024pi_0,intelligence2025pi_,bjorck2025gr00t};
Section~\ref{sec:prelim-state} surveys these designs. Because each design
ships inside a different system, reported numbers confound the state
interface with pretraining, backbone, data, and evaluation protocol. No
prior study moves the same state signal across interfaces, history
depths, and injection routes with everything else fixed; that controlled
comparison is the role of this paper.

\paragraph{Temporal context in imitation learning.}
Observation histories reduce partial observability and support
multi-stage control, with recurrent modules, memory tokens, and
compressed latents as alternatives to raw frame stacks
\citep{bulatov2022recurrent,dai2026robomme,shi2025memoryvla,li2026spatial}.
The same
histories, however, invite shortcuts: behavioral cloning can copy its own
recent trajectory instead of attending to the scene
\citep{de2019causal,wen2020fighting}. A depth sweep alone cannot separate
temporal content from the extra conditioning slots that come with it, so
we pair every history model with a repeat-current control that fixes slot
count and interface while removing temporal change.

\paragraph{Empirical studies of VLA components.}
Recent work isolates individual VLA design choices through controlled
ablation, covering latent-action supervision, geometric features, and
memory interfaces \citep{lin2026pixels,yang2026understanding,dai2026robomme}.
We bring this methodology to proprioception and hold to one discipline
throughout: every claim about a design rests on independently trained
systems compared under one protocol, so that a difference in success rate
can be attributed to the state design rather than to a perturbation of a
single checkpoint.

\section{Preliminaries}

\subsection{Vision--Language--Action Policies}
\label{sec:prelim-policy}

This paper studies VLA policies that are jointly conditioned on vision,
language, and robot proprioceptive state. At decision time $t$, the robot
receives a visual observation $o_t \in \mathcal{O}$, a language instruction
$\ell \in \mathcal{L}$, and an ordered state window over the most recent $K$
steps,
\begin{equation}
  S_t^{(K)} = \left(s_{t-K+1}, \ldots, s_t\right), \qquad
  s_t \in \mathbb{R}^{d_s},
\end{equation}
where $s_t$ describes the robot's current kinematic configuration, such as
end-effector pose, base pose, and gripper state; $K{=}1$ means that only the
current step is used. Conditioned on these inputs, the policy $\pi_\theta$
generates a continuous action chunk of length $H$,
\begin{equation}
  \hat{\mathbf{a}}_{t:t+H-1} =
  \pi_\theta\!\left(o_t, \ell, S_t^{(K)}\right) \in \mathbb{R}^{H \times d_a},
\end{equation}
where $d_a$ is the per-step action dimension. The model is trained on a
dataset of demonstration trajectories
$\mathcal{D} = \{(o_t^{(i)}, \ell^{(i)}, S_t^{(i,K)},
\mathbf{a}_{t:t+H-1}^{(i)})\}$ with an action-generation objective
$\mathcal{L}_{\mathrm{act}}$ that matches the predicted action chunk to the
expert action sequence. Crucially, state interfaces do not differ in this
prediction target; they differ in how the state window $S_t^{(K)}$ is
represented and in how it is injected into the vision--language backbone or
the action-generation module.

\subsection{Robot Proprioception and Its Use in VLAs}
\label{sec:prelim-state}

Robot proprioception---joint angles, end-effector pose, gripper state, and
base or whole-body configuration---is the only policy input that lives in
the same continuous physical space as the output actions, and current VLAs
wire it in strikingly different ways: $\pi_{0.5}$ serializes it into
discrete text tokens in the VLM prompt \citep{intelligence2025pi_},
OpenVLA-OFT projects it continuously into the language-model sequence
\citep{kim2025fine}, and GR00T~N1 feeds a state embedding directly to the
action head \citep{bjorck2025gr00t}. Around these designs, a rapidly growing
line of work explores what else state can do: as alignment supervision or a
contrastive regularizer for representation learning
\citep{wen2025rosa,DBLP:journals/corr/abs-2510-01711,DBLP:journals/corr/abs-2601-04061},
as tokenized input that actively guides visual reasoning or routes a
dedicated expert stream
\citep{DBLP:journals/corr/abs-2602-06575,DBLP:journals/corr/abs-2601-14133},
as a normalized representation for cross-embodiment and whole-body control
\citep{davies2025tenma,DBLP:journals/corr/abs-2604-07993}, as a continuously
evolving signal
for long-horizon manipulation and memory
\citep{DBLP:conf/aaai/QiYCBFCW26,DBLP:journals/corr/abs-2512-20166,shi2025memoryvla},
and as a cue for estimating execution risk and detecting false completion
\citep{zhai2026cofreevla,DBLP:journals/corr/abs-2601-16667}. At the same
time, several studies
warn in the opposite direction: naively fused state can dominate vision,
letting a policy ``complete'' a task from its internal progress while
ignoring visual failure \citep{DBLP:journals/corr/abs-2601-16667}, can
suppress visual learning around motion-phase transitions
\citep{DBLP:journals/corr/abs-2602-12032}, and can serve as an
action-correlation shortcut in behavioral cloning
\citep{de2019causal,wen2020fighting}. The community has thus explored many
ways of using state, and the evidence pulls in both directions---but these
findings come from different backbones, data, and protocols, which is
precisely why this paper compares state interfaces, history depths, and
injection routes under a single controlled scaffold.

\section{Methodology}

\subsection{Problem Setup}
\label{sec:backbone}

We adopt $\pi_{0.5}$ \citep{intelligence2025pi_}---a VLM coupled with a
flow-matching action expert \citep{flowmatching}---as the base policy for
the entire study, instantiating the policy class of
Section~\ref{sec:prelim-policy}. Proprioceptive state can enter this
architecture at different points and in different forms; holding everything
else fixed, we turn the three questions posed in the introduction into three
controlled variables:
the \emph{representation} of state (discrete or continuous, and through
which interface), the \emph{history length} $K$, and the \emph{injection
site} (the VLM side or the action side). All five interfaces share the same
action-generation objective, data pipeline, and training recipe; they differ
only in how the same state window is represented and where it enters the
policy---the VLM prefix, the action prefix, a dedicated state stream, or
feature modulation. Cross-interface comparisons therefore evaluate complete
state-conditioning systems, rather than an abstract ``injection site''
detached from capacity and computation path; the task-family partition and
the evaluation protocol appear in Section~\ref{sec:protocol}.

\subsection{Proprioceptive State Representation}
\label{sec:staterep}

Each raw state frame $s_q \in \mathbb{R}^{16}$ contains the end-effector
position and quaternion in the base frame, the mobile-base position
and quaternion in the world frame, and two gripper joint positions. We keep the shorthand
``proprioceptive state'' below, but the base pose also carries world-frame
localization, so its effect cannot be read as purely internal motor feedback.

Except for the discrete state prompt, which uses only the current frame, all
continuous interfaces receive state sequences with identical numerical
content, temporal order, and history length. Each frame is passed
independently through a two-layer projector to form one continuous state
token,
\begin{equation}
  h_i = \phi_2\!\left(\mathrm{Swish}\!\left(\phi_1(s_i)\right)\right),
\end{equation}
where $\phi_1$ lifts the state input, zero-padded from 16 to the framework's
fixed 32 dimensions, to width $d$, and $\phi_2$ preserves that width, with
$d{=}2048$ for the VLM prefix and $d{=}1024$ for the action prefix, the
state expert, and feature modulation, matching the hidden width of the host
module in each case. Sharing the per-frame projector guarantees a consistent
encoding process across history depths, but it does not equalize the total
parameter count or compute across interfaces; these differences are
quantified in Figure~\ref{fig:cost}.

\subsection{State Interfaces}
\label{sec:interfaces}

\paragraph{State Prompt (\eef{}).}
This interface reuses the native text entry of $\pi_{0.5}$: each dimension
of the current state is quantized into 256 bins and serialized through the
existing tokenizer, producing roughly 66 prompt tokens that join the
language instruction in the VLM input. Because the bins map onto the
pretrained vocabulary, this route adds no trainable parameters and is the
only interface whose state tokens pass through the same embedding space as
language; by construction it supports only the current frame.

\paragraph{VLM Prefix (\vlm{}).}
Per-frame state tokens ($d{=}2048$) are inserted into the bidirectional
VLM prefix, after the image and language tokens. State first participates
in multimodal context modeling---every image and language token can attend
to it---and then influences action generation indirectly through the
conditioning prefix.

\paragraph{Action Prefix (\ac{}).}
State tokens ($d{=}1024$) are placed in the causal action suffix, ahead of
the noisy action tokens, so that they participate directly in the action
expert's velocity-field prediction at every denoising step without first
being compressed into the VLM representation. This is the most direct
route from state to action.

\paragraph{State Expert (\se{}).}
A dedicated state-processing stream is added alongside the VLM and the
action expert, giving state its own sequence-modeling path that exchanges
information with the action module during generation. This is the
largest-capacity design: state is neither compressed into the VLM nor
folded into the action suffix, but processed by its own transformer
stack.

\paragraph{Feature Modulation (\sm{}).}
State is kept as a separate conditioning memory rather than as ordinary
sequence tokens. Each layer of the action expert reads it through
cross-attention and predicts a per-feature scale $\gamma$ and shift
$\beta$ that continually modulate the action features:
\begin{equation}
  \mathrm{Mod}(z; S) = \left(1 + \gamma(z, S)\right) \odot z + \beta(z, S).
\end{equation}

Under single-frame conditioning, \eef{}, \vlm{}, \ac{}, \se{}, and \sm{}
add 0, 4.26M, 1.08M, 199.30M, and 123.84M trainable parameters,
respectively. \se{} and \sm{} thus change both topology and capacity, so
the five interfaces jointly answer which \emph{complete system} is
effective under a given task structure; the closest route pair (\vlm{}
versus \ac{}) is examined in Section~\ref{sec:route}.

\section{Experiments}

Our experiments systematically study how proprioceptive state should be used
in a unified flow-matching VLA framework, organized around three core
questions.
\textbf{RQ1 (state utility and interface):} does explicitly conditioning on
the current proprioceptive state improve closed-loop control, and how do the
gains of different state interfaces change with the control task?
\textbf{RQ2 (temporal depth):} how many frames of raw state history provide a
real performance benefit, and does the improvement genuinely come from
temporal variation in state rather than from additional conditioning slots?
\textbf{RQ3 (injection route):} should state enter the vision--language
backbone or the action-generation module, and does this route preference
change with history depth?
To answer these questions, we compare a no-state baseline, five state
interfaces, and history configurations from 1 to 96 frames within the fixed
$\pi_{0.5}$ framework, evaluated closed-loop on 45 atomic tasks and 20
in-distribution composite tasks. The design separates two levels of claim:
comparisons between independently trained systems assess the utility of a
complete interface, whereas the slot-matched true-history versus
repeat-current control isolates the training benefit of genuine temporal
variation from that of added conditioning capacity. Fixed-checkpoint probes
that trace how state reaches action generation are reported in
Appendix~A.

\subsection{Experimental Protocol}
\label{sec:protocol}

\paragraph{Benchmark.}
We evaluate the effect of proprioceptive state interfaces and state history
on closed-loop robot control on RoboCasa365
\citep{nasiriany2026robocasa365}, which builds on the RoboCasa
simulation platform and contains diverse kitchen scenes, robot initial
configurations, object instances, and manipulation tasks. All methods use the
same visual inputs, language instructions, action space, and evaluation
program; visual observations consist of three views from a left exterior
camera, a right exterior camera, and a wrist camera.

\paragraph{Atomic benchmark.}
Atomic tasks assess single-stage control. We pre-partition the
RoboCasa365 atomic tasks by their dominant manipulation semantics---not
post hoc by model performance---into three families: A (rearrangement
and pick-and-place, probing large-range positioning of the end effector
and mobile base), B (articulated-object interaction, probing sustained
contact and motion-phase modeling), and C (knob, switch, and appliance
control, where small workspaces demand high local precision). Navigation
is out of scope. Each family contains 15 representative tasks (45 in
total) and trains a separate category expert, so cross-family results
reflect interface behavior under different control demands rather than
one unified multi-task policy. Every task is evaluated with 50
closed-loop rollouts under a fixed, identical task list and episode-seed
schedule.

\paragraph{Composite benchmark.}
Composite evaluation uses the
\texttt{lifelong\_learning\_phase2} setting: 20 task types, each chaining
two or three atomic subgoals within one episode. Each interface trains
one policy jointly on all 20 types; evaluation holds out only new
episodes, so Phase~2 measures in-distribution composite control rather
than generalization to unseen semantics. In addition to the default
end-effector-pose state representation, we repeat the key composite-task
comparisons using joint-angle state under the identical training and
evaluation protocol. The action representation remains end-effector
deltas in both settings, isolating the effect of the state coordinate
system. Every composite task is evaluated with 25 closed-loop rollouts
under the same fixed seed schedule.

\paragraph{Experimental notes.}
The primary metric is closed-loop task success rate (SR). A rollout counts as
successful only if the environment-defined success condition is satisfied
within the horizon; for Phase~2, success requires completing all constituent
subgoals in an episode. All models are initialized from the same pretrained
$\pi_{0.5}$ checkpoint and fully fine-tuned on the corresponding task data
under a shared optimization recipe, keeping the data pipeline, action
representation, learning-rate schedule, and training budget matched across
interfaces; \se{} and \sm{} train with a slightly lower nominal sample
exposure due to hardware allocation, so we do not use them for
capacity-matched claims. Random seeds are fixed before training and
evaluation, and the same seed
schedule is reused for all models within a benchmark to control scene
initialization, object instances, object placements, and other environment
randomness. Slot-matched temporal comparisons also fix images, language,
state-slot count, expert actions, and initial flow noise. Paired
task-bootstrap intervals resample aligned tasks for two fixed checkpoints;
they measure evaluation and task-sampling uncertainty rather than
optimization variance, and the interface-by-depth sweep is exploratory and
unadjusted for multiplicity.

\subsection{Results and Analysis}

\subsubsection{RQ1: Does Current State Improve Control?}

\paragraph{Conditioning on the current state is beneficial overall.}
Table~\ref{tab:state-utility} shows that on the 45 atomic tasks the
no-state model reaches a macro success rate of 54.6\%, and the point
estimates of all five state interfaces exceed this baseline, with gains
ranging from $+1.1$ (\ac{}) to $+3.1$ points (\eef{}). In particular,
\eef{} reaches 57.7\% with a paired task-bootstrap 95\% interval of
$[0.2, 6.1]$, providing direct statistical evidence that the current
state can improve closed-loop control; the intervals of the remaining
interfaces include zero, so we read their gains as a consistent positive
tendency rather than individually supported effects.

\paragraph{There is no task-agnostic best interface.}
Beneath the macro average, the family-level rankings reverse. Family~A
rearrangement tasks favor \eef{} (68.7\%, $+7.0$ over no-state), while
the continuous interfaces gain far less there ($-0.1$ to $+2.6$).
Family~B articulation tasks reverse this order: \vlm{} leads at 68.8\%
($+6.1$), with \se{} and \sm{} close behind ($+5.8$ and $+5.5$), whereas
\eef{} drops to mid-pack ($+1.6$). Family~C knob-and-switch control is
both the hardest family (no-state 39.5\%) and the most selective: \se{}
leads at 42.8\% ($+3.3$), and \vlm{} is the only interface that falls
below the baseline ($-1.2$). Each interface thus has a family where it
shines and a family where it adds little or even hurts---a pattern
consistent with the three families imposing different control demands,
and one that a single benchmark-wide average would completely hide.

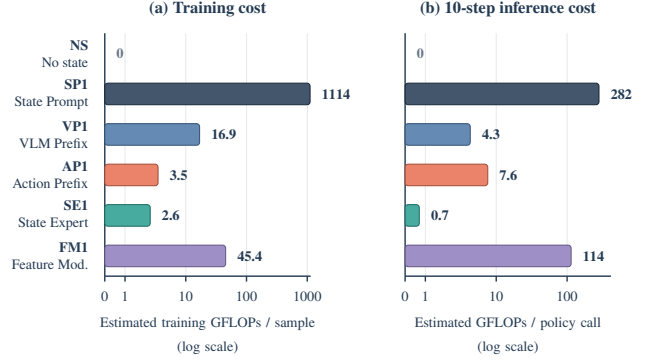
\begin{figure}[t]
\centering
\begingroup
\definecolor{ccSP}{HTML}{243B53}
\definecolor{ccVP}{HTML}{4C78A8}
\definecolor{ccAP}{HTML}{E76F51}
\definecolor{ccSE}{HTML}{2A9D8F}
\definecolor{ccFM}{HTML}{8E7DBE}
\definecolor{ccGrid}{HTML}{E5E7EB}
\definecolor{ccAxis}{HTML}{243B53}
\definecolor{ccMuted}{HTML}{6B7785}
\resizebox{\columnwidth}{!}{%
\begin{tikzpicture}[
  x=1cm,y=1cm,
  bottomlabel/.style={
    anchor=north,
    align=center,
    text=ccAxis,
    font=\tiny,
    inner sep=0pt,
    text width=3.20cm,
    execute at begin node={\setlength{\baselineskip}{7.2pt}}
  }
]
\def\rowdy{0.60}      
\def\barh{0.16}       
\def\ulen{0.30}       
\def\xB{4.45}
\foreach \i/\idlab/\namelab in {0/NS/No state, 1/SP1/State Prompt, 2/VP1/VLM Prefix, 3/AP1/Action Prefix, 4/SE1/State Expert, 5/FM1/Feature Mod.}{
  \node[anchor=east, align=right, font=\tiny, text=ccAxis] at (-0.12, -\i*\rowdy)
    {\textbf{\idlab}\\ \namelab};
}
\foreach \k in {0,1,2,3}{
  \draw[ccGrid!95, line width=0.40pt]
    ({0.30+0.90*\k}, 0.30) -- ({0.30+0.90*\k}, {-5*\rowdy-0.30});
}
\draw[ccAxis, line width=0.6pt] (0, {-5*\rowdy-0.30}) -- (3.05, {-5*\rowdy-0.30});
\draw[ccAxis, line width=0.6pt] (0, 0.30) -- (0, {-5*\rowdy-0.30});
\foreach \k/\lab in {0/0, 1/1, 2/10, 3/100, 4/1000}{
  \pgfmathsetmacro\tx{\k==0 ? 0 : 0.30+0.90*(\k-1)}
  \draw[ccAxis, line width=0.6pt] (\tx, {-5*\rowdy-0.30}) -- (\tx, {-5*\rowdy-0.38});
  \node[font=\tiny, anchor=north, text=ccAxis] at (\tx, {-5*\rowdy-0.40}) {\lab};
}
\node[bottomlabel] at (1.525, {-5*\rowdy-0.90})
  {Estimated training GFLOPs / sample\\[3.0pt](log scale)};
\node[font=\scriptsize\bfseries, anchor=south, text=ccAxis] at (1.525, 0.42)
  {(a) Training cost};
\foreach \k in {0,1,2}{
  \draw[ccGrid!95, line width=0.40pt]
    ({\xB+0.30+1.05*\k}, 0.30) -- ({\xB+0.30+1.05*\k}, {-5*\rowdy-0.30});
}
\draw[ccAxis, line width=0.6pt] (\xB, {-5*\rowdy-0.30}) -- ({\xB+3.05}, {-5*\rowdy-0.30});
\draw[ccAxis, line width=0.6pt] (\xB, 0.30) -- (\xB, {-5*\rowdy-0.30});
\foreach \k/\lab in {0/0, 1/1, 2/10, 3/100}{
  \pgfmathsetmacro\tx{\k==0 ? \xB : \xB+0.30+1.05*(\k-1)}
  \draw[ccAxis, line width=0.6pt] (\tx, {-5*\rowdy-0.30}) -- (\tx, {-5*\rowdy-0.38});
  \node[font=\tiny, anchor=north, text=ccAxis] at (\tx, {-5*\rowdy-0.40}) {\lab};
}
\node[bottomlabel] at ({\xB+1.525}, {-5*\rowdy-0.90})
  {Estimated GFLOPs / policy call\\[3.0pt](log scale)};
\node[font=\scriptsize\bfseries, anchor=south, text=ccAxis] at ({\xB+1.525}, 0.42)
  {(b) 10-step inference cost};
\foreach \i/\col/\v/\vl in {1/ccSP/1114/1114, 2/ccVP/16.9/16.9, 3/ccAP/3.5/3.5, 4/ccSE/2.6/2.6, 5/ccFM/45.4/45.4}{
  \pgfmathsetmacro\blen{\v < 1 ? 0.30*\v : 0.30 + 0.90*log10(\v)}
  \draw[rounded corners=.45mm, draw=\col!68!black, fill=\col!86, line width=0.25pt]
    (0, {-\i*\rowdy-\barh}) rectangle (\blen, {-\i*\rowdy+\barh});
  \node[font=\tiny\bfseries, anchor=west, text=ccAxis] at ({\blen+0.05}, {-\i*\rowdy}) {\vl};
}
\node[font=\tiny\bfseries, anchor=west, text=ccMuted] at (0.05, 0) {0};
\foreach \i/\col/\v/\vl in {1/ccSP/282/282, 2/ccVP/4.3/4.3, 3/ccAP/7.6/7.6, 4/ccSE/0.7/0.7, 5/ccFM/114/114}{
  \pgfmathsetmacro\blen{\v < 1 ? 0.30*\v : 0.30 + 1.05*log10(\v)}
  \draw[rounded corners=.45mm, draw=\col!68!black, fill=\col!86, line width=0.25pt]
    (\xB, {-\i*\rowdy-\barh}) rectangle ({\xB+\blen}, {-\i*\rowdy+\barh});
  \node[font=\tiny\bfseries, anchor=west, text=ccAxis] at ({\xB+\blen+0.05}, {-\i*\rowdy}) {\vl};
}
\node[font=\tiny\bfseries, anchor=west, text=ccMuted] at ({\xB+0.05}, 0) {0};
\end{tikzpicture}%
}
\endgroup
\caption{Estimated marginal compute of the single-frame interfaces (log
scale). (a) Training-time marginal GFLOPs per sample; (b) ten-step inference
marginal GFLOPs per policy call.}
\label{fig:cost}
\end{figure}

\paragraph{The interfaces differ sharply in computational cost.}
Figure~\ref{fig:cost} (Appendix~C) reports the theoretical marginal compute relative
to the no-state model. \eef{} serializes the state into roughly 66
discrete prompt tokens, adding about 1114 training GFLOPs per sample and
282 GFLOPs per ten-step policy call---the most expensive design by two
orders of magnitude on the training side. The continuous interfaces are
far cheaper: \vlm{} adds 16.9/4.3 (training/inference), \ac{} 3.5/7.6,
and \se{} only 2.6/0.7, with \sm{} in between at 45.4/114. Weighing
performance against compute, \eef{} offers the clearest overall benefit
but at the highest price, while \se{} and \sm{} reach nearly the same
macro point estimate (57.6\%) at a small fraction of the marginal
compute---a relevant trade-off when the state interface must scale with
history length.

\begin{table*}[t]
\centering
\caption{Success rates (\%) of single-frame state-interface systems on 45
atomic tasks. \textbf{Bold} indicates the best result and
\underline{underline} the second best; ties are preserved. Green and red
numbers in parentheses show improvement and degradation over the matched
no-state baseline, respectively. $\dagger$ marks a paired task-bootstrap
95\% confidence interval that excludes zero.}
\label{tab:state-utility}
\begingroup
\newcommand{\tblgain}[1]{\textcolor{green!55!black}{\scriptsize $(\uparrow\,#1)$}}
\newcommand{\tbldrop}[1]{\textcolor{red!70!black}{\scriptsize $(\downarrow\,#1)$}}
\newcommand{\tblsame}{\textcolor{gray}{\scriptsize $(\pm\,0.0)$}}
\setlength{\tabcolsep}{4.5pt}
\renewcommand{\arraystretch}{1.12}
\small
\begin{tabular*}{\textwidth}{@{\extracolsep{\fill}}lrrrrrr@{}}
\toprule
\textsc{Task}
& \shortstack{\textsc{NS}\\[-1pt]\scriptsize No state}
& \shortstack{\textsc{SP1}\\[-1pt]\scriptsize State Prompt}
& \shortstack{\textsc{VP1}\\[-1pt]\scriptsize VLM Prefix}
& \shortstack{\textsc{AP1}\\[-1pt]\scriptsize Action Prefix}
& \shortstack{\textsc{SE1}\\[-1pt]\scriptsize State Expert}
& \shortstack{\textsc{FM1}\\[-1pt]\scriptsize Feature Mod.} \\
\midrule
\textbf{Atomic A}
& 61.7
& \textbf{68.7} \tblgain{7.0}
& 63.2 \tblgain{1.5}
& 61.7 \tblsame
& 61.6 \tbldrop{0.1}
& \underline{64.3} \tblgain{2.6} \\
\textbf{Atomic B}
& 62.7
& 64.3 \tblgain{1.6}
& \textbf{68.8} \tblgain{6.1}
& 65.9 \tblgain{3.2}
& \underline{68.5} \tblgain{5.8}
& 68.2 \tblgain{5.5} \\
\textbf{Atomic C}
& 39.5
& \underline{40.3} \tblgain{0.8}
& 38.3 \tbldrop{1.2}
& 39.6 \tblgain{0.1}
& \textbf{42.8} \tblgain{3.3}
& \underline{40.3} \tblgain{0.8} \\
\midrule
\rowcolor{blue!7}
\textbf{Atomic macro}
& 54.6
& \textbf{57.7} \tblgain{3.1}$^{\dagger}$
& 56.8 \tblgain{2.1}
& 55.7 \tblgain{1.1}
& \underline{57.6} \tblgain{3.0}
& \underline{57.6} \tblgain{2.9} \\
\bottomrule
\end{tabular*}
\endgroup

\end{table*}

\subsubsection{RQ2: Is State History Useful, and How Much Is Needed?}

\begin{figure*}[t]
\centering
\includegraphics[width=0.98\textwidth]{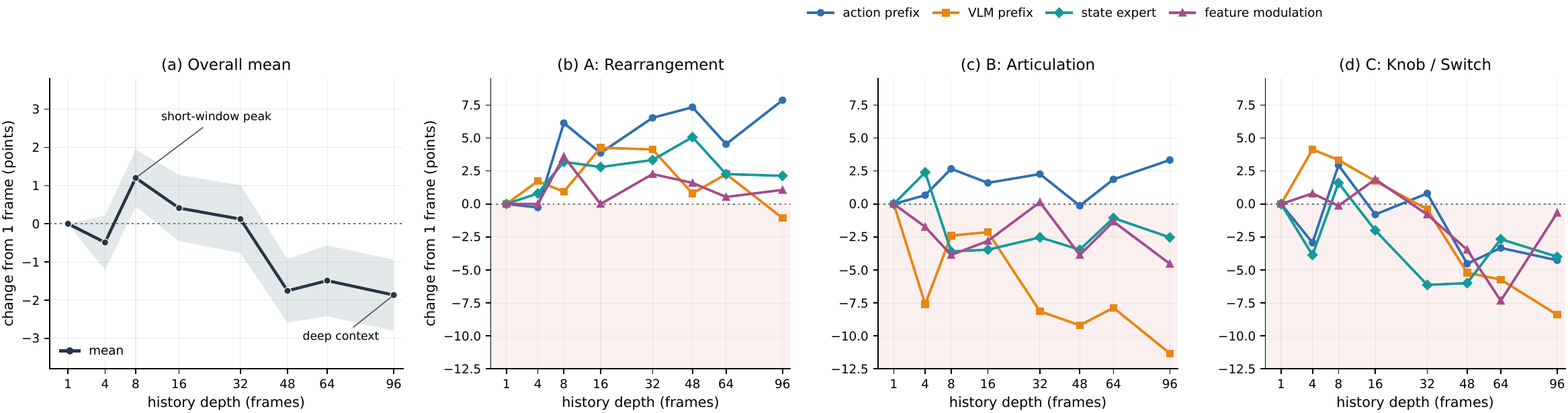}
\caption{Effect of state-history depth on the RoboCasa atomic subset.
(a) Mean change relative to the matched single-frame model, revealing a
bounded short-history benefit rather than monotonic improvement with depth.
(b--d) Results by task family: families A and B are relatively robust, whereas
long uncompressed histories substantially degrade performance on family C,
especially when injected through the VLM prefix.}
\label{fig:depth}
\end{figure*}

\paragraph{Short state history helps, whereas long raw history can be harmful.}
We first study history depth on the RoboCasa atomic subset, where the shorter
task horizon allows the effect of temporal context to be isolated more
directly. Figure~\ref{fig:depth} shows a clear non-monotonic trend: short
histories improve performance over the corresponding single-frame models,
whereas deeper uncompressed histories provide no additional benefit and
eventually degrade control. The effect is not uniform across task families.
Families A and B are comparatively tolerant to increasing context, while
family C suffers a pronounced drop under long histories. This suggests that
raw temporal context is useful only within a limited horizon; beyond that
range, redundant or stale state information can interfere with action
prediction, particularly in tasks that require more precise state-to-action
alignment.

\begin{figure*}[!t]
\centering
\begingroup
\footnotesize
\definecolor{dtSP}{HTML}{243B53}
\definecolor{dtVP}{HTML}{4C78A8}
\definecolor{dtAP}{HTML}{E76F51}
\definecolor{dtSE}{HTML}{2A9D8F}
\definecolor{dtFM}{HTML}{8E7DBE}
\definecolor{dtGrid}{HTML}{E5E7EB}
\definecolor{dtMuted}{HTML}{6B7785}
\begin{tikzpicture}[
  x=1.22cm, y=1cm,
  every node/.style={inner sep=1pt},
  barap/.style={rounded corners=.35mm, fill=dtAP!86, draw=dtAP!68!black, line width=0.26pt},
  barvp/.style={rounded corners=.35mm, fill=dtVP!86, draw=dtVP!68!black, line width=0.26pt},
  barse/.style={rounded corners=.35mm, fill=dtSE!84, draw=dtSE!68!black, line width=0.26pt},
  barfm/.style={rounded corners=.35mm, fill=dtFM!84, draw=dtFM!68!black, line width=0.26pt},
  barsp/.style={rounded corners=.35mm, fill=dtSP!92, draw=dtSP!78!black, line width=0.26pt},
  vlab/.style={font=\tiny, anchor=south},
]
\begin{scope}
  \foreach \v/\yy in {52/0.60, 54/1.20, 56/1.80, 58/2.40, 60/3.00} {
    \draw[dtGrid, very thin] (0.08,\yy) -- (4.95,\yy);
    \node[anchor=east, text=dtMuted, font=\tiny] at (0.04,\yy) {\v};
  }
  \draw[dtMuted, line width=0.45pt] (0.08,0) -- (0.08,3.00);
  \draw[dtMuted, densely dashed, line width=0.4pt] (0.08,1.38) -- (4.95,1.38);
  \fill[barap] (0.30,0) rectangle (0.66,1.71); \node[vlab] at (0.48,1.73) {55.7};
  \fill[barvp] (0.76,0) rectangle (1.12,2.04); \node[vlab] at (0.94,2.06) {56.8};
  \fill[barse] (1.22,0) rectangle (1.58,2.28); \node[vlab] at (1.40,2.30) {57.6};
  \fill[barfm] (1.68,0) rectangle (2.04,2.28); \node[vlab] at (1.86,2.42) {57.6};
  \fill[barsp] (2.14,0) rectangle (2.50,2.31);
    \node[vlab, text=dtSP!92!black] at (2.35,2.56) {\textbf{57.7}$^{\dagger}$};
    \node[font=\tiny, text=dtSP, anchor=south] at (2.32,2.80) {$\star$};
  \fill[barap] (3.00,0) rectangle (3.36,2.88);
    \node[vlab, text=dtAP!72!black] at (3.18,2.90) {\textbf{59.6}};
    \node[font=\tiny, text=dtAP, anchor=south] at (3.18,3.14) {$\star$};
  \fill[barvp] (3.46,0) rectangle (3.82,2.22); \node[vlab] at (3.64,2.24) {57.4};
  \fill[barse] (3.92,0) rectangle (4.28,2.40); \node[vlab] at (4.10,2.54) {58.0};
  \fill[barfm] (4.38,0) rectangle (4.74,2.22); \node[vlab] at (4.56,2.24) {57.4};
  \node[text=dtMuted, font=\tiny, inner sep=1pt] at (2.75,1.38) {54.6};
  \draw[dtMuted, line width=0.45pt] (0.08,0) -- (4.95,0);
  \node[anchor=north, font=\scriptsize] at (1.40,-0.06) {$K{=}1$};
  \node[anchor=north, font=\scriptsize] at (3.87,-0.06) {$K{=}8$};
  \node[anchor=south, font=\scriptsize\bfseries, text=dtSP!92!black] at (2.51,3.30) {Atomic (45 tasks)};
\end{scope}
\begin{scope}[xshift=7.26cm]
  \foreach \v/\yy in {24/0.60, 28/1.20, 32/1.80, 36/2.40, 40/3.00} {
    \draw[dtGrid, very thin] (0.08,\yy) -- (4.50,\yy);
    \node[anchor=east, text=dtMuted, font=\tiny] at (0.04,\yy) {\v};
  }
  \draw[dtMuted, line width=0.45pt] (0.08,0) -- (0.08,3.00);
  \draw[dtMuted, densely dashed, line width=0.4pt] (0.08,1.26) -- (4.50,1.26);
  \fill[barap] (0.30,0) rectangle (0.66,1.23); \node[vlab] at (0.48,1.25) {28.2};
  \fill[barvp] (0.76,0) rectangle (1.12,2.16);
    \node[vlab, text=dtVP!72!black] at (0.94,2.18) {\textbf{34.4}};
    \node[font=\tiny, text=dtVP, anchor=south] at (0.94,2.42) {$\star$};
  \fill[barse] (1.22,0) rectangle (1.58,0.87); \node[vlab] at (1.40,0.89) {25.8};
  \fill[barfm] (1.68,0) rectangle (2.04,1.17); \node[vlab] at (1.86,1.19) {27.8};
  \fill[barap] (2.54,0) rectangle (2.90,2.85);
    \node[vlab, text=dtAP!72!black] at (2.72,2.87) {\textbf{39.0}};
    \node[font=\tiny, text=dtAP, anchor=south] at (2.72,3.11) {$\star$};
  \fill[barvp] (3.00,0) rectangle (3.36,2.07); \node[vlab] at (3.18,2.09) {33.8};
  \fill[barse] (3.46,0) rectangle (3.82,1.20); \node[vlab] at (3.64,1.22) {28.0};
  \fill[barfm] (3.92,0) rectangle (4.28,1.83); \node[vlab] at (4.10,1.85) {32.2};
  \node[text=dtMuted, font=\tiny, inner sep=1pt] at (2.29,1.26) {28.4};
  \draw[dtMuted, line width=0.45pt] (0.08,0) -- (4.50,0);
  \node[anchor=north, font=\scriptsize] at (1.17,-0.06) {$K{=}1$};
  \node[anchor=north, font=\scriptsize] at (3.41,-0.06) {$K{=}8$};
  \node[anchor=south, font=\scriptsize\bfseries, text=dtSP!92!black] at (2.29,3.30)
    {Composite, EEF-pose state (20 tasks)};
\end{scope}
\begin{scope}[xshift=13.97cm]
  \foreach \v/\yy in {24/0.60, 28/1.20, 32/1.80, 36/2.40, 40/3.00} {
    \draw[dtGrid, very thin] (0.08,\yy) -- (2.65,\yy);
    \node[anchor=east, text=dtMuted, font=\tiny] at (0.04,\yy) {\v};
  }
  \draw[dtMuted, line width=0.45pt] (0.08,0) -- (0.08,3.00);
  \fill[barap] (0.30,0) rectangle (0.66,1.71); \node[vlab] at (0.48,1.73) {31.4};
  \fill[barvp] (0.76,0) rectangle (1.12,2.04);
    \node[vlab, text=dtVP!72!black] at (0.94,2.06) {\textbf{33.6}};
    \node[font=\tiny, text=dtVP, anchor=south] at (0.94,2.30) {$\star$};
  \fill[barap] (1.62,0) rectangle (1.98,2.43);
    \node[vlab, text=dtAP!72!black] at (1.72,2.45) {\textbf{36.2}};
  \fill[barvp] (2.08,0) rectangle (2.44,2.37); \node[vlab] at (2.34,2.39) {35.8};
  \node[font=\tiny, text=dtMuted, anchor=south] at (2.03,2.72) {$\approx$};
  \draw[dtMuted, line width=0.45pt] (0.08,0) -- (2.65,0);
  \node[anchor=north, font=\scriptsize] at (0.71,-0.06) {$K{=}1$};
  \node[anchor=north, font=\scriptsize] at (2.03,-0.06) {$K{=}8$};
  \node[anchor=south, font=\scriptsize\bfseries, text=dtSP!92!black] at (1.36,3.30)
    {Composite, joint-angle state};
\end{scope}
\begin{scope}[yshift=-0.73cm, xshift=1.45cm]
  \fill[dtAP!86] (0.10,0) rectangle (0.32,0.14);
  \draw[dtAP!68!black, line width=0.26pt] (0.10,0) rectangle (0.32,0.14);
  \node[anchor=west, font=\tiny] at (0.38,0.07) {\ac{}};

  \fill[dtVP!86] (2.00,0) rectangle (2.22,0.14);
  \draw[dtVP!68!black, line width=0.26pt] (2.00,0) rectangle (2.22,0.14);
  \node[anchor=west, font=\tiny] at (2.28,0.07) {\vlm{}};

  \fill[dtSE!84] (3.90,0) rectangle (4.12,0.14);
  \draw[dtSE!68!black, line width=0.26pt] (3.90,0) rectangle (4.12,0.14);
  \node[anchor=west, font=\tiny] at (4.18,0.07) {\se{}};

  \fill[dtFM!84] (5.80,0) rectangle (6.02,0.14);
  \draw[dtFM!68!black, line width=0.26pt] (5.80,0) rectangle (6.02,0.14);
  \node[anchor=west, font=\tiny] at (6.08,0.07) {\sm{}};

  \fill[dtSP!92] (7.70,0) rectangle (7.92,0.14);
  \draw[dtSP!78!black, line width=0.26pt] (7.70,0) rectangle (7.92,0.14);
  \node[anchor=west, font=\tiny] at (7.98,0.07) {\eef{} (single-frame only)};

  \node[anchor=west, font=\tiny, text=dtMuted] at (2.05,-0.25)
    {dashed: \nostate{} baseline \qquad $\star$: per-cluster best point estimate};
\end{scope}
\end{tikzpicture}
\endgroup
\caption{Where and when proprioceptive state helps, and where it should
enter. Macro success rate (\%); dashed lines mark the \nostate{} baseline
(54.6 atomic, 28.4 composite). \emph{Left}: on near-Markovian atomic tasks
every interface stays within $+1.1$ to $+5.0$ points of the baseline
($\dagger$: the only paired 95\% CI that excludes zero), and the largest
change from adding history is again on the action side (\ac{}
$55.7\!\rightarrow\!59.6$). \emph{Middle}: on composite tasks the preferred
entry point switches with the temporal budget --- with a single frame only
the VLM-side prefix (\vlm{}) clearly exceeds the baseline, while with a
short history the action-side prefix (\ac{}) becomes the strongest entry.
\emph{Right}: the same direction replicates under a joint-angle state; at
matched $K{=}8$ the \ac{}/\vlm{} difference lies within the paired
task-bootstrap noise band ($\approx$). Across all three panels the
$K{=}1$ best is a VLM-side entry (\eef{}/\vlm{}) and the $K{=}8$ best is
\ac{}. Slot-matched controls attribute the composite \ac{} history gain to
genuine temporal variation (Table~\ref{tab:temporal}); all other entries
are single-seed point estimates.}
\label{fig:crossover-bars}
\end{figure*}
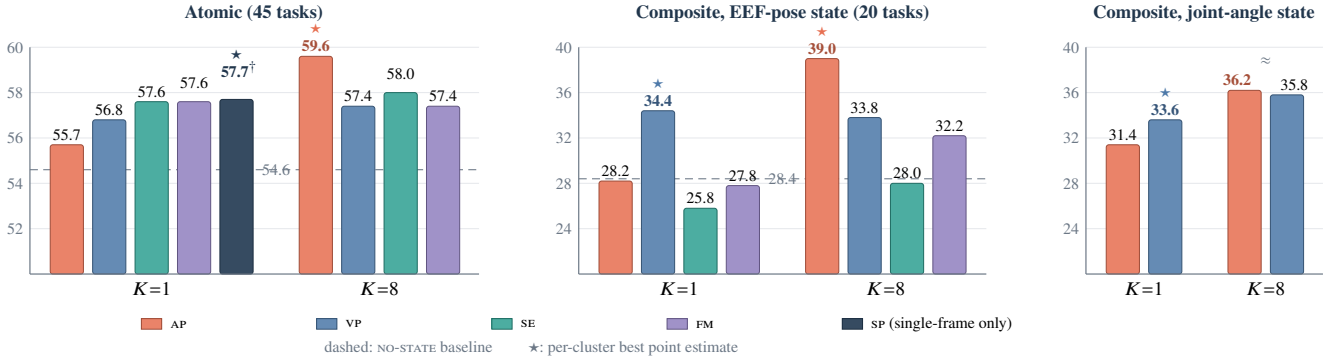

\paragraph{An eight-frame history provides a practical operating point.}
Across the atomic-task sweep, the strongest and most consistent gains occur
within the short-history regime, with (K{=}8) providing a favorable balance
between temporal information and context redundancy. We therefore adopt
eight frames as the default temporal-conditioning recipe for the subsequent
experiments. This choice is empirical rather than universal: the key finding
is not that eight frames are optimal for every task, but that compact histories
are consistently more reliable than long raw sequences.

\paragraph{Composite tasks confirm that the benefit transfers to long-horizon
control.}
We next evaluate the same recipe on the 20 RoboCasa Phase-2 composite tasks,
which involve longer horizons and more complex multi-stage behavior.
Table~\ref{tab:temporal} shows that extending the state input from a single
frame to an eight-frame history improves the overall performance of most
conditioning interfaces, with the clearest benefit appearing on the action
side. In contrast, injecting the same history through the VLM prefix remains
less consistent. The atomic-subset observation therefore transfers to the
more challenging composite setting: recent state evolution is useful, but it
is most effectively exploited when it directly conditions action generation.

\begin{table}[!t]
\centering
\caption{Composite-task temporal comparisons. The first two blocks report
exploratory complete-system changes from $K{=}1$ to $K{=}8$ for the default
EEF-pose state and for a joint-angle state representation trained and
evaluated under the identical protocol (the action space stays EEF deltas);
all entries are single-seed point estimates. The final row is the
slot-matched temporal-content test on the EEF-state \ac{} policy:
\emph{current-only} copies the current state into eight \ac{} slots, whereas
\emph{genuine history} supplies eight ordered state frames. SR and
$\Delta$SR are percentage points.}
\label{tab:temporal}
\begingroup
\newcommand{\tblpos}[1]{\textcolor{green!55!black}{#1}}
\newcommand{\tblneg}[1]{\textcolor{red!70!black}{#1}}
\small
\setlength{\tabcolsep}{3.8pt}
\renewcommand{\arraystretch}{1.08}
\begin{tabular*}{\columnwidth}{@{\extracolsep{\fill}}lrrr@{}}
\toprule
Interface / control & $K{=}1$ & $K{=}8$ & $\Delta$SR \\
\midrule
\multicolumn{4}{@{}l}{\emph{EEF-pose state (default)}} \\
AP & 28.2 & 39.0 & \tblpos{+10.8} \\
VP & 34.4 & 33.8 & \tblneg{$-0.6$} \\
SE & 25.8 & 28.0 & \tblpos{+2.2} \\
FM & 27.8 & 32.2 & \tblpos{+4.4} \\
\midrule
\multicolumn{4}{@{}l}{\emph{Joint-angle state (same protocol)}} \\
AP & 31.4 & 36.2 & \tblpos{+4.8} \\
VP & 33.6 & 35.8 & \tblpos{+2.2} \\
\midrule
\multicolumn{4}{@{}l}{\emph{Slot-matched control (EEF-pose state)}} \\
AP: current-only $\rightarrow$ genuine history
& 30.8 & 39.0 & \tblpos{+8.2} \\
\bottomrule
\end{tabular*}
\endgroup
\end{table}

\paragraph{The gain reflects temporal structure rather than additional tokens.}
Increasing the history length also increases the number of conditioning
tokens, creating a potential capacity confound. To control for this factor, we
compare the eight-frame model with a slot-matched variant that repeats the
current state across the same number of positions. The repeated-state control
falls substantially short of the genuinely ordered history, and the paired
task-bootstrap confidence interval excludes zero (Appendix~B). Thus, the improvement
cannot be explained by additional conditioning slots alone; it arises from
the temporal variation encoded across successive states.

\paragraph{The conclusion is robust to the state representation.}
Finally, we repeat the comparison using joint-angle states instead of
end-effector states, while keeping the action space and training protocol
unchanged. Both representations preserve the same qualitative advantage of
short history, and their matched differences remain within the uncertainty of
the benchmark. The usefulness of temporal conditioning is therefore not tied
to a particular coordinate system. Relative to history depth and injection
location, the choice between joint and end-effector states is a secondary
design factor.

Taken together, the experiments establish a consistent two-stage result.
The atomic-task analysis identifies a bounded temporal regime in which short
state histories are beneficial and long raw histories are detrimental,
especially for family C. Using the resulting (K{=}8) recipe, the composite-task
experiments further show that the benefit persists under longer-horizon
control, survives token-count matching, and holds across both joint-space and
end-effector state representations. These findings support compact
action-side state history as a robust default for temporal conditioning.

\subsubsection{RQ3: Where Should State Enter?}
\label{sec:route}

\paragraph{With a single frame ($K{=}1$), state favors the VLM side.}
On composite tasks, which demand stage tracking over longer horizons,
VLM-side injection is the clear single-frame winner: \vlm{}$_1$ reaches
34.4\% versus 28.2\% for \ac{}$_1$ with the EEF-pose state, and 33.6\%
versus 31.4\% with joint angles, while every other entry stays within a
few points of the 28.4\% no-state baseline. On the atomic suite the
single-frame leaders sit within 0.1 points of one another, so the
preference there is not resolved---the VLM-side advantage emerges
precisely where a single observation helps contextualize the
visual--language representation for multi-stage control.

\paragraph{With a short history ($K{=}8$), the advantage shifts decisively
to the action side.}
The $K{=}1\!\rightarrow\!8$ gains are largest when the sequence directly
conditions the action head: $+10.8$ points with the EEF-pose state, $+4.8$
with joint angles, and $+3.9$ on the atomic suite, whereas the same
history in the VLM prefix yields only $-0.6$, $+2.2$, and $+0.6$. The
endpoints tell the same story: with eight frames, \ac{} holds the best
entry in every panel (59.6\% atomic, 39.0\% composite) after being the
weakest or near-weakest single-frame interface. Together with the
slot-matched control in RQ2, the two findings compose into a simple
design rule for our setting: \emph{inject single-frame state into the
VLM, but route multi-frame state history to the action head}.

\paragraph{Scope of the routing rule.}
The crossover persists across both state representations and both task
suites, and fixed-checkpoint probes trace it to a difference in data flow:
VLM-side state measurably alters the multimodal context before action
generation, whereas action-side state acts directly inside the action
expert (Appendix~A). The rule nonetheless remains a directional pattern
rather than an established causal mechanism or a statistical ranking:
under joint angles the two routes converge at $K{=}8$ (36.2\% versus
35.8\%), the routes differ in projection width and computation path, and
most comparisons rely on a single training seed.

\section{Conclusion}

We presented a controlled study of how a flow-matching VLA should use
proprioceptive state, decomposing an otherwise ad hoc wiring decision
into three measurable design axes: representation, temporal depth, and
injection route. Three findings summarize the evidence. Current state
improves closed-loop control selectively: the discrete prompt yields the
only interval-supported macro gain, and the best interface changes with
the task family. State history has a bounded useful range: short
histories help, long raw histories hurt, and the slot-matched control
attributes the gain to genuine temporal content rather than added
conditioning capacity. The preferred injection route depends on the
temporal budget: a single frame is most effective in the VLM prefix,
whereas a short history is most effective when it directly conditions the
action expert. These findings compose into a concrete design
default---inject the current frame into the VLM, route short histories to
the action head, and validate anything deeper---and the underlying
evaluation procedure is directly reusable for auditing future state-aware
VLAs. Two limitations remain. Our experiments lack real-robot validation,
and the state studied here is purely kinematic, leaving out force,
tactile, and other sensing modalities; extending the protocol to physical
platforms and to multimodal state is our next step.

\clearpage
\bibliography{refs}

\clearpage
\appendix
\section*{Appendix}
\label{app:supplement}

This appendix provides supporting material in three parts. Part A presents
fixed-checkpoint probes that trace how proprioceptive state reaches action
generation. Part B uses a paired composite-task case study to localize the
benefit of ordered short history. Part C gives the analytic derivation of the
marginal training and inference costs reported in the main paper.

\subsection*{A. Probe Experiments}
\label{app:probes}

The main experiments reveal a routing crossover: a current state frame is
most effective through a VLM-side interface, whereas an ordered short history
is used most reliably through the action prefix. We connect these system-level
results to three matched fixed-checkpoint probes. The probes trace where state
first changes the computation and how that change reaches the generated
action; they complement, rather than replace, closed-loop evaluation.

\subsubsection*{A.1 Matched State Interventions}

For each of the 45 atomic tasks, we cache four expert-action contexts and
evaluate the same image, instruction, action target, and flow-noise sample under
\emph{true-state} and \emph{state-off} forward passes, giving 180 matched
contexts per interface. Continuous state is disabled by zeroing its projected
embedding while preserving token positions and masks. For the discrete state
prompt, we use both a zero-state serialization and a masked state span.

For VLM-prefix probes, language-to-image attention redistribution is the
total-variation distance between normalized image-patch distributions under
true state and state-off. Image-token representation change is the relative
$\ell_2$ distance between their hidden states. For action-flow probes, let
\begin{equation}
  c_t=\hat a^{\mathrm{true}}_t-\hat a^{\mathrm{off}}_t,
  \qquad
  r_t=a^\star-\hat a^{\mathrm{off}}_t ,
\end{equation}
where $c_t$ is the state-conditioned correction at Euler step $t$ and $r_t$
is the residual from state-off to the expert action. We report correction
direction $\cos(c_t,r_t)$ and normalized magnitude
$\|c_t\|_2/\|r_t\|_2$.

\subsubsection*{A.2 Current State Has an Early VLM-Side Path}

Across the final six VLM layers, the 45-task mean language-to-image attention
redistribution is 17.3\% for \vlm{}$_1$ and 22.0\% for \vlm{}$_8$; the
corresponding image-token relative $\ell_2$ changes are 19.6\% and 26.2\%.
Both responses grow toward later layers. Disabling \ac{}$_8$ leaves these
VLM-prefix quantities unchanged because its state tokens enter only after
the prefix has been formed. Thus, VLM-side state has an early
contextualization route, while action-prefix state must influence generation
directly inside the action expert.

\subsubsection*{A.3 State-Conditioned Correction Accumulates During Generation}

Every interface produces a correction that grows as noise is transformed
into an action, but route and history depth determine how strongly the
correction aligns with the expert residual. With one continuous state frame,
the final alignment is 0.245 for \vlm{}$_1$ and 0.079 for \ac{}$_1$;
normalized magnitudes are 0.297 and 0.174. Ordered short history changes the
action-side response: moving from \ac{}$_1$ to \ac{}$_8$ raises final
alignment from 0.079 to 0.270 and normalized magnitude from 0.174 to 0.382.
Across the 45 paired tasks, the eight-minus-one increases are $+0.191$ for
alignment and $+0.208$ for magnitude, with task-bootstrap 95\% intervals
$[+0.143,+0.239]$ and $[+0.171,+0.244]$.

\subsubsection*{A.4 Representative Action-Conditioned Spatial Attention}

The VLM-prefix state visibly redistributes the image regions read by action
queries. The one-frame action prefix changes little spatially, while ordered
short history increases its downstream response without rewriting the VLM
prefix. This example is illustrative; the quantitative layer-wise and
flow-trajectory results above aggregate all 45 tasks. Together, the probes
support a data-flow interpretation of the main result: current VLM-side state
can alter the multimodal context before action generation, whereas temporal
action-side state acts directly through the action expert.

\subsection*{B. Case Study}
\label{app:case-study}

The aggregate experiments establish that compact state histories can improve
composite control, particularly through the action-prefix route. We use
\textsc{PrepareToast} to identify where this advantage appears within an
episode and when a fixed policy responds to genuine temporal variation.

\subsubsection*{B.1 Task and Paired Protocol}

\textsc{PrepareToast} requires the policy to place two task-relevant items
and then return to the cabinet to close it. We compare independently trained
one-frame and eight-frame action-prefix policies, \ac{}$_1$ and
\ac{}$_8$, using 50 paired episode seeds. Pairing fixes the scene, object
instances, object placements, and simulator randomization. Four monotonic
milestones are derived from native task predicates and summarized in
Table~\ref{tab:case-milestones}.

\begin{table}[b]
\centering
\caption{\textbf{Native \textsc{PrepareToast} milestones.} Each predicate is
monotonic and defined independently of either policy's outcome.}
\label{tab:case-milestones}
\small
\setlength{\tabcolsep}{4pt}
\renewcommand{\arraystretch}{1.08}
\begin{tabular}{@{}lp{0.73\columnwidth}@{}}
\toprule
Stage & Native predicate \\
\midrule
S1 & The first task-relevant item has been placed. \\
S2 & Both task-relevant items have been placed. \\
S3 & The cabinet has been reclosed. \\
S4 & The gripper has been released after completion. \\
\bottomrule
\end{tabular}
\end{table}

Within the fixed \ac{}$_8$ checkpoint, we also compute a non-executed
counterfactual action after replacing the ordered eight-frame history by
eight copies of the current state. Images, language, parameters, state-slot
count, and initial flow noise are held fixed. The resulting distance
\begin{equation}
  D_q =
  \left\|
  \hat{\mathbf{a}}^{\,\mathrm{true}}_q -
  \hat{\mathbf{a}}^{\,\mathrm{repeat}}_q
  \right\|_2
\end{equation}
measures fixed-policy sensitivity to temporal variation rather than to the
number of state slots.

\subsubsection*{B.2 Where the Advantage Appears}

The two policies progress similarly through the early placement stages:
\ac{}$_1$ reaches S1 and S2 in 90\% and 64\% of episodes, while
\ac{}$_8$ reaches them in 96\% and 68\%. The separation begins after both
items have been placed. \ac{}$_1$ reaches the cabinet-reclosed milestone S3
in 30\% of episodes, whereas \ac{}$_8$ reaches it in 56\%. The same
$+26$-point difference persists through S4 and final success, with a paired
episode-bootstrap interval of $[+10,+42]$ points. Conditional on reaching
S2, completion of S3 rises from 46.9\% to 82.4\%.

\subsubsection*{B.3 The Policy Reads Genuine Temporal Variation}

The mean true-history versus repeat-current action distance is 0.198 within a
stage, 0.361 before a boundary, 1.033 at a boundary, and 0.748
afterward. Boundary sensitivity is therefore 5.2 times the within-stage
value. Because every paired forward pass shares the same image, instruction,
state-slot count, and flow noise, this contrast cannot be attributed to more
tokens or a different visual trajectory. The two analyses jointly localize
the benefit to the late placement-to-closure transition and show that the
trained policy is most sensitive to ordered state variation around progress
boundaries. Their co-occurrence is descriptive evidence of temporal use, not
a causal mediation claim.

\subsubsection*{B.4 Case-Study Takeaway}

The paired analysis separates two questions that an aggregate success rate
alone cannot resolve. The independently trained \ac{}$_1$--\ac{}$_8$
comparison measures system-level temporal utility: the performance
separation appears only after the two placement subgoals have been completed
and control must return to the cabinet. The fixed-checkpoint
true-history--repeat-current comparison measures temporal reliance: the
eight-frame policy changes its action most strongly near the native progress
boundaries while images, parameters, flow noise, and state-slot count remain
matched. The two observations therefore agree at the behavioral level:
ordered recent evolution is read most strongly where the active subgoal
changes, and the trained history policy is more likely to complete the
corresponding late transition. We retain the distinction between association
and mediation; the case study localizes a consistent use pattern but does not
claim that boundary sensitivity is the sole cause of the success gain.

Several design choices make this localization auditable. The four milestones
are monotonic predicates supplied by the task definition, rather than stages
inferred from either policy's actions or final outcome. Paired episode seeds
hold the scene and simulator randomization fixed for the system-level
comparison, so each reach-rate difference refers to the same set of task
instances. The fixed-policy intervention then removes a different
alternative explanation: repeating the current state preserves the number
and positions of all state tokens, so the measured action change is tied to
their temporal content rather than additional conditioning capacity. These
two controls operate at complementary levels. The first asks whether a
policy trained with recent history progresses farther through the same
episodes; the second asks whether that trained policy changes its prediction
when only the order-dependent state content is removed.

\begin{figure*}[p]
\centering
\includegraphics[width=0.88\textwidth]
{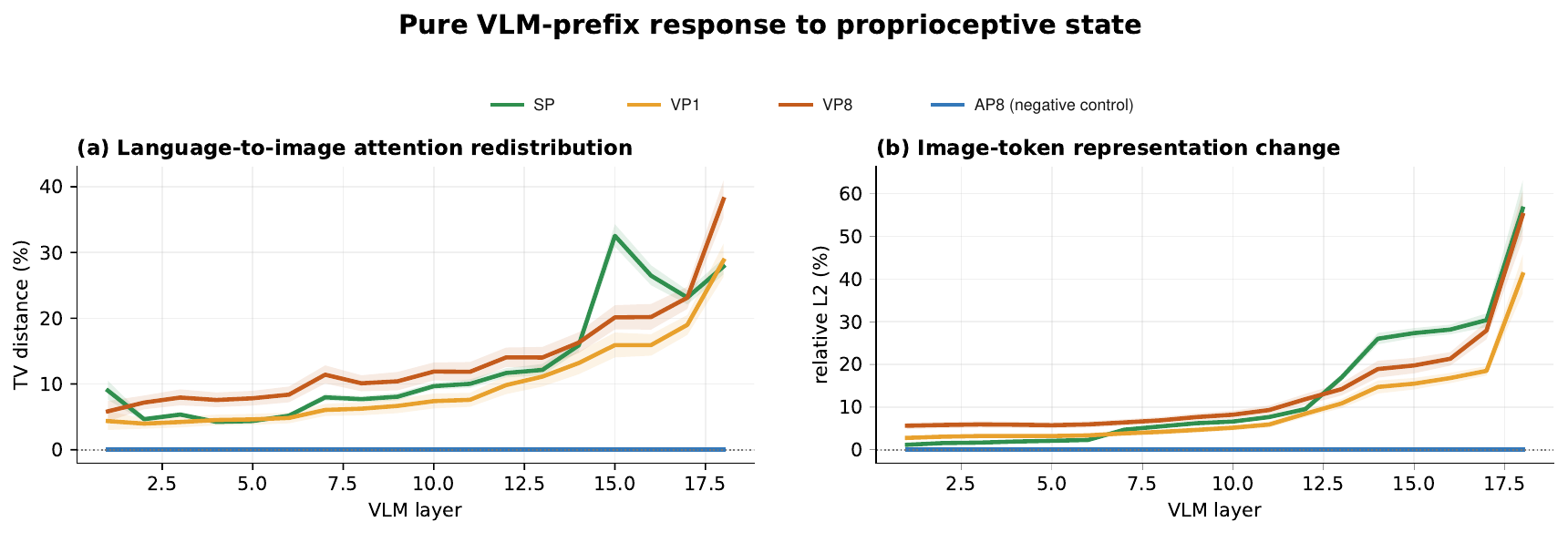}
\caption{\textbf{VLM-prefix response to proprioceptive state.}
Layer-wise language-to-image attention redistribution and image-token
representation change are measured between matched true-state and state-off
passes. VLM-side state progressively alters the shared multimodal prefix,
whereas action-prefix state leaves that prefix unchanged and acts downstream
through action generation.}
\label{fig:prefix-probe}

\vspace{0.6em}

\includegraphics[width=0.92\textwidth]
{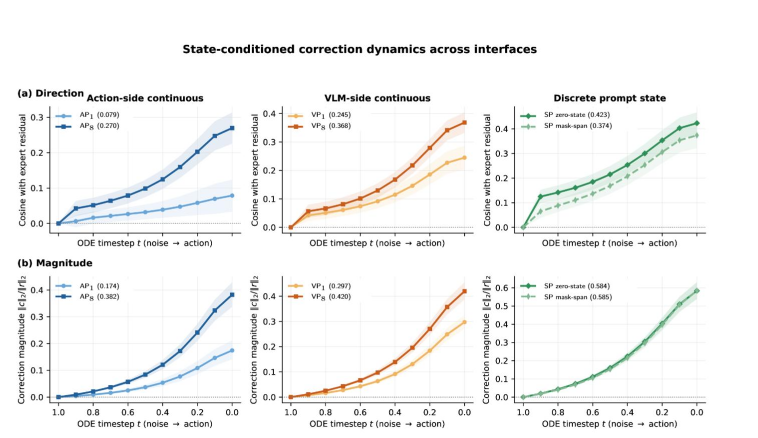}
\caption{\textbf{State-conditioned correction over the ten-step flow
trajectory.}
The top row reports alignment between the true-minus-off correction and the
state-off-to-expert residual. The bottom row reports correction magnitude
normalized by that residual. Columns compare action-side continuous state,
VLM-side continuous state, and discrete prompt state. Curves average 45
tasks; bands are task-bootstrap 95\% intervals.}
\label{fig:flow-probe}
\end{figure*}

\begin{figure*}[t!]
\centering
\includegraphics[width=0.66\textwidth]
{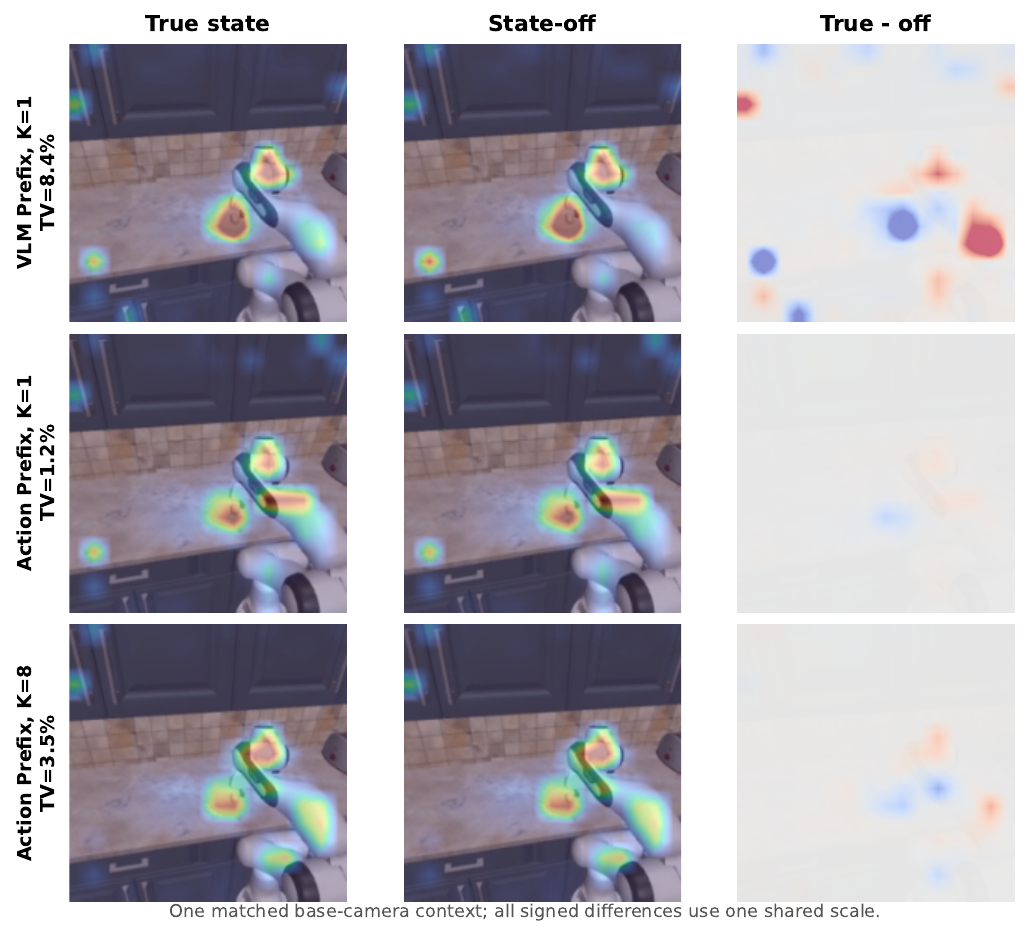}
\caption{\textbf{Representative action-conditioned spatial attention.}
Matched true-state and state-off passes show how a one-frame VLM prefix and
one- and eight-frame action prefixes redistribute action-conditioned visual
attention. The visualization is descriptive: it reveals where state changes
the spatial readout, while the flow probe in Figure~\ref{fig:flow-probe}
measures how that change reaches action generation.}
\label{fig:action-attention-probe}

\vspace{0.45em}

\centering
\includegraphics[width=0.92\textwidth]
{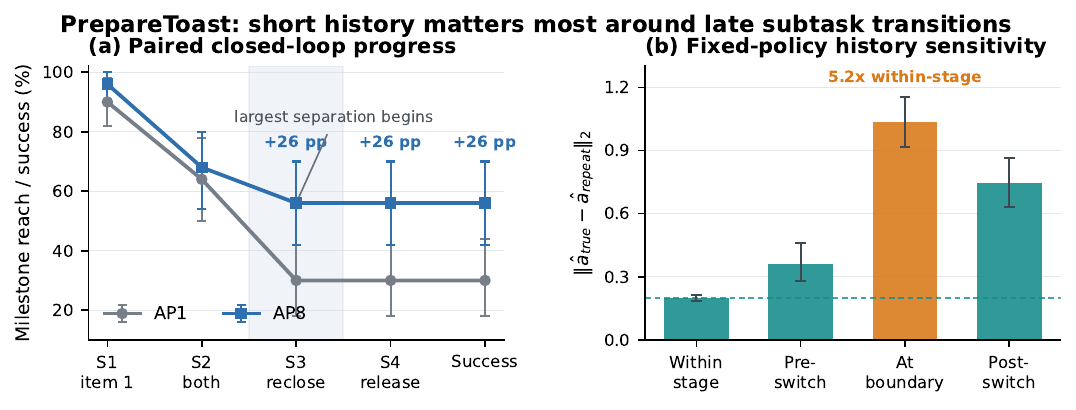}
\caption{\textbf{Paired \textsc{PrepareToast} case study.}
(a) AP$_1$ and AP$_8$ are similar through placement of the two items
(S1--S2), after which their reach rates separate when control must return
to closing the cabinet (S3). AP$_8$ improves S3, S4, and final success by
26 percentage points over 50 paired episode seeds.
(b) Within the fixed AP$_8$ checkpoint, replacing ordered history with
repeated current state produces the largest action change near native
milestone boundaries. Error bars are episode-bootstrap 95\% intervals.}
\label{fig:prepare-toast-case}
\end{figure*}

\FloatBarrier

\subsection*{C. Detailed Compute-Cost Derivation}
\label{app:compute}

This section expands the cost analysis summarized in the main paper. The
purpose is to compare the \emph{semantic marginal arithmetic} introduced by
each state interface relative to a token-free no-state policy. Consequently,
the calculation charges State Prompt for its 66 serialized state tokens even
when an implementation pads both state and no-state prompts to the same
static tensor shape. The estimates are hardware independent and should not
be interpreted as wall-clock latency or peak memory.

\subsubsection*{C.1 Shared Backbone and Counting Convention}

The analytic model follows the evaluated $\pi_{0.5}$ scaffold. It contains
$L=18$ Transformer layers, three image streams with 256 tokens each, 16
instruction tokens, and $A=50$ action tokens. Thus, before state is added,
the conditioning prefix has
\begin{equation}
  P = 3\times256 + 16 = 784
\end{equation}
active tokens. The VLM width is $d_v=2048$ with an FFN width of 16384; the
action expert width is $d_a=1024$ with an FFN width of 4096. The grouped
attention inner width is $H=8\times256=2048$, and policy inference uses
$N=10$ Euler denoising steps.

We count one multiply-add as two floating-point operations. The marginal
forward cost of processing one additional token through all Transformer
layers is approximated by
\begin{equation}
 F_{\mathrm{tok}}(d,m)
 = 2L\!\left[d(H+2\!\times\!256+H)+3dm\right],
\end{equation}
where $d$ and $m$ are the host module's hidden and FFN widths. A shared
two-layer state projector, $32\!\rightarrow\!d\!\rightarrow\!d$, contributes
\begin{equation}
 F_{\mathrm{enc}}(d,K)=2K(32d+d^2)
\end{equation}
for $K$ state frames. Because the same projector is reused frame by frame,
its parameter count is independent of history depth.

\subsubsection*{C.2 Attention Expansion by Injection Route}

Adding $K$ bidirectional state tokens to the VLM prefix enlarges prefix and
prefix-to-action attention by
\begin{equation}
 \Delta F_{\mathrm{VP,attn}}
 =4HL\left(2PK+K^2+AK\right).
\end{equation}
Placing the same tokens in the causal action suffix instead gives
\begin{equation}
 \Delta F_{\mathrm{AP,attn}}
 =4HL\left(PK+\frac{K(K+1)}{2}+AK\right).
\end{equation}
When a state token has already been encoded and cached in the prefix, each
Euler step pays an additional suffix-to-prefix attention cost
\begin{equation}
 \Delta F_{\mathrm{cache}}=4HLAK.
\end{equation}
These expressions are added to the corresponding state-encoder and
per-token Transformer costs. State Prompt uses the VLM-prefix expression
with 66 serialized tokens and no additional learned parameters. Feature
Modulation uses no ordinary sequence tokens; its cost instead comprises one
state-encoder pass plus per-layer cross-attention and scale/shift projection.

\subsubsection*{C.3 Training and Ten-Step Inference}

The reported training column approximates executed arithmetic as
\begin{equation}
 \Delta F_{\mathrm{train}}
 \simeq 4\,\Delta F_{\mathrm{forward}},
\end{equation}
because the Transformer blocks are rematerialized during backpropagation.
This factor accounts for the forward evaluation and the recomputation needed
by the backward pass; it is an arithmetic convention, not a measured timing
multiplier. During inference, VLM-prefix state is encoded once and reused
through the ten denoising steps, whereas action-prefix processing and
feature-modulation operations inside the action expert are repeated at each
step. This difference explains why Action Prefix has low marginal training
cost but a larger ten-step inference increment than VLM Prefix.

\subsubsection*{C.4 Interpretation}

The derivation separates three notions that are easy to conflate. Active
token count determines sequence-level arithmetic, extra parameters describe
learned capacity, and measured throughput additionally depends on padding,
compilation, rematerialization, and hardware scheduling. State Prompt, for
example, adds no parameters but is analytically expensive because 66 semantic
tokens traverse the VLM. Its near-baseline measured throughput remains
compatible with this estimate because the implementation uses a fixed padded
prompt shape. Conversely, State Expert and Feature Modulation add substantial
learned capacity without adding many ordinary sequence tokens.

Temporal depth must likewise be considered jointly with injection route.
VLM-prefix training cost grows quickly because every additional state token
participates in the large VLM prefix, whereas action-prefix inference repeats
state-token processing across all denoising steps. The analysis therefore
does not identify a universally cheapest interface; it makes explicit which
resource each route consumes and provides the counting basis for the main
paper's performance--efficiency comparison.

\end{document}